# Core Box Image Recognition and its Improvement with a New Augmentation Technique

Evgeny E. Baraboshkin[a,b], Andrey E. Demidov[b], Denis M. Orlov[a,b], Dmitry A. Koroteev[a,b]

[a]Skolkovo Institute of Science and Technology, [b]Digital Petroleum

Bolshoy Boulevard 30, bld. 1, Moscow, Russia

**Corresponding author:**

Evgeny E. Baraboshkin, evgenii.baraboshkin@skoltech.ru

**Present/permanent address**

Skolkovo Institute of Science and Technology, Bolshoy Boulevard 30, bld. 1, Moscow, Russia 121205

**Abstract**

Most methods for automated full-bore rock core image analysis (description, colour, properties distribution, etc.) are based on separate core column analyses. The core is usually imaged in a box because of the significant amount of time taken to get an image for each core column. The work presents an innovative method and algorithm for core columns extraction from core boxes. The conditions for core boxes' imaging may differ tremendously. Such differences are disastrous for machine learning algorithms which need a large dataset describing all possible data variations. Still, such images have some standard features – a box and core. Thus, we can emulate different environments with a unique augmentation described in this work. It is called template-like augmentation (TLA). The method is described and tested on various environments, and results are compared on an algorithm trained on both "traditional" data and a mix of traditional and TLA data. The algorithm trained with TLA data provides better metrics and can detect core on most new images, unlike the algorithm trained on data without TLA. The algorithm for core column extraction implemented in an automated core description system speeds up the core box processing by a factor of 20.

**Keywords: Core Box Image; Segmentation; Convolutional Neural Networks; Geology; Template-like Augmentation; Core Column Extraction; Machine Vision**

## 1. Introduction

Various approaches have been developed to study full-bore core images by a computer due to rapid quantity and quality data increase. The images may be used for automated rock typing (Baraboshkin et al., 2020; Lepistö, 2005; Patel et al., 2017; Thomas et al., 2011), (Alzubaidi et al., 2021) and different properties distribution analysis (Egorov, 2019; Khasanov et al., 2016; Prince and Chitale, 2008; Wieling, 2013). Most of them are based on an analysis of a separated core column, usually stored in a box. Some are based on a statistical analysis of a separate core column image, and others are based on machine learning and deep learning algorithms. As core column images are usually stored in a box, these columns should be extracted from the full-bore core box image first (figure 1) to proceed with image-based rock typing. A special algorithm should be developed to perform this extraction. There is no available research on the core column extraction from an image.

This work presents a method and approach to extract a core column from a core box image and discuss challenges and solutions which may appear during algorithm development.

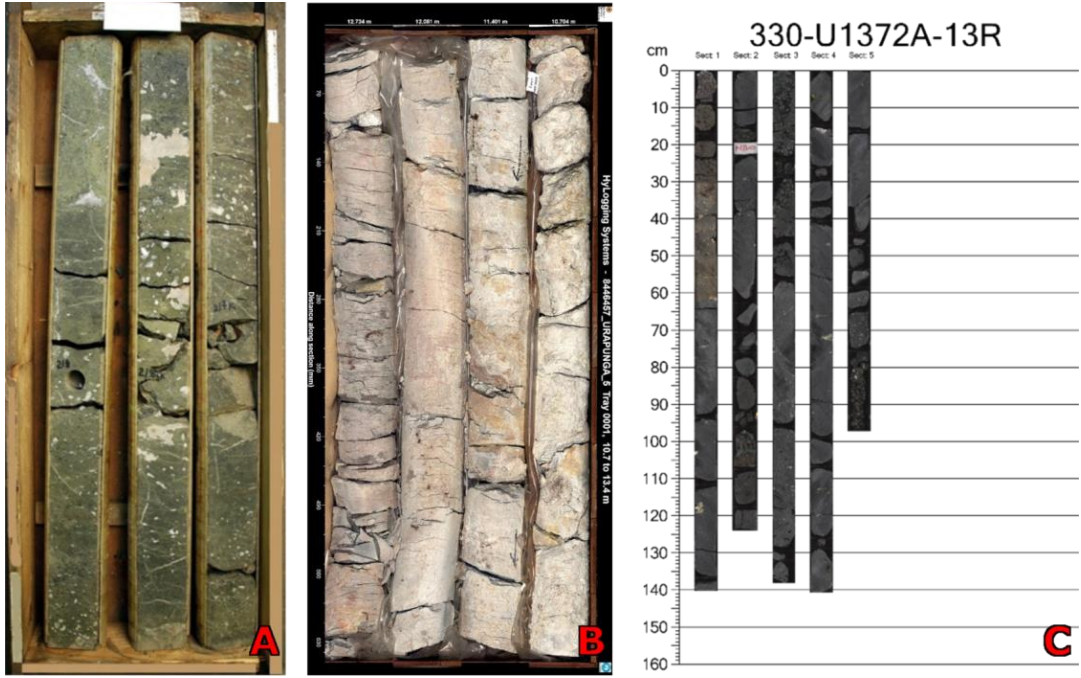

Figure 1. Different data types received from national and international data repositories: NOPIMS (National Offshore Petroleum Information Management System (B), IODP (Integrated Ocean Drilling Program) (C), RosGeolfond (Russian federal geological fund) (A).

Images in figure 1 show that each sample image was taken in various conditions (lightning, camera position, different tools inside the frame area etc.). Different approaches can be applied to extract information from the core box. An untrained person can, in most cases, separate the core and the box on the image manually, but the computer vision algorithms are not that consistent. The significant issues of computer vision are the following. Traditional computer vision methods will work with specific data types only and should be tuned to each new dataset (or even an image within the dataset) separately. If a researcher wants to implement a universal algorithm, he/she can develop a complicated decision-making system based on computer vision or train a machine learning algorithm. One can also use a deep learning algorithm. Deep learning methods (which can be used in the computer vision system) will not work from time to time due to domain adaptation problems. It is hard to create such an algorithm that will work the same way on different data types due to a great variety of such data, especially with insufficient training data.

One of the commonly used ways to interpret an image in computer vision is semantic segmentation with supervised deep learning. Many examples of labelled data are required for such an approach as one needs to describe data variety sufficiently. This is an elusive target for a specific task with great diversity, even if one creates own dataset. Thus, the domain adaptation problem related to the unrepresented data appears. This issue is commonly resolved by complex algorithms that demand much time and computational resources (like generative adversarial networks).(Gong et al., 2018; Shorten and Khoshgoftaar, 2019).

Current work presents a method and a practical implementation of domain adaptation problem elimination with a lightweight solution. We assume that our approach can be applied in machine vision systems with template-like data, where parts of the image are similar (like surveillance systems, flaw detection, store shelves packing check). This will help solve the domain adaptation problem in research and production model development. The method is driven by deep augmentation of an image with easy-to-implement ways based on initial labels.

During the development of the method, it was proven to be effective on core box image segmentation tasks with low accessibility of training examples and a great variety of data. The primary purpose was to get stable results and decrease false-positive responses on a wide range of data that may not be presented in the initial dataset.

## 2. Materials and Methods

This research implemented a template-like data augmentation (TLA), described in section 2.3 below, to enhance the algorithm performance.

Code was written in Python (version 3.7) (Van Rossum and Drake, 2011). Different Python libraries were used:

- OpenCV (4.1.1 (Bradski, 2000))
- Numpy (1.18.1 (Travis, 2006))
- Augmentation library Albumentation (0.4.1 (Buslaev et al., 2018)).
- The Segmentation Models library (1.0.0 (Yakubovskiy, 2019)) was used to test a hypothesis based on the use of convolutional neural networks (CNN) applied for segmentation.
- Keras library was used to build CNN's architecture (2.3.1 (Chollet et al., 2015)) with Tensorflow backend (1.15) (Martín Abadi, Ashish Agarwal, Paul Barham et al., 2015) and Pytorch (1.3 (Ketkar, 2017)) with the use of one Nvidia 1080TI GPU instance.
- Matplotlib (3.1.1 (Hunter, 2007)) was used for plotting the results. Scikit-Learn (0.22, (Pedregosa et al., 2011)) used for model tests.

### 2.1. Data description

The database included different types of core boxes (figure 2). A core is a sample of rock extracted from the Earth by drilling and stored in separate containers (or boxes). The length of a core was from 0.5 to 1 meter. The diameter was about 0.1 m. As a part of the description, the core boxes are shot with a high-resolution camera (figure 1) (the image size is up to 4000x6000px) under daylight (100%) and, sometimes, under ultraviolet light (43%).

The following characteristics are typical for core boxes:

1) a core may destruct during extraction and reallocation and has a different shape.

2) separate core parts may be removed from the box for experiments.

3) the initial size of the image may vary on a large scale.

This work aims to separate core columns from the background part of the image like core box, colour pallet and foreign objects – the best suitable algorithm for such a task (bearing in mind the characteristics above) is the image segmentation algorithm. A dataset should be prepared to train a segmentation algorithm consisting of pixel-accuracy labelled images. Such a dataset was created.

The dataset included 1010 manually labeled images of different resolution (minimal size 2860x860px ~ 100 dpi recalculated to core diameter equal to 10 cm, maximum size – 6000x4000px ~ 300 dpi). It was separated into training (734 images) and test (276) sets to test the algorithm.

The labels were manually added to an image mask in the Gimp (GNU Image Manipulation Program) application. The mask was stored in a grayscale format, which has 255 values of grey (a regular uint-8 image). We used 0 for the background and 255 for the core (figure 2). This information about key (the core) and value is stored in JSON-format file (Bray, 2014) and looks like that: {labels: {" core_column": 255} ...}. The image and mask are stored in different folders.

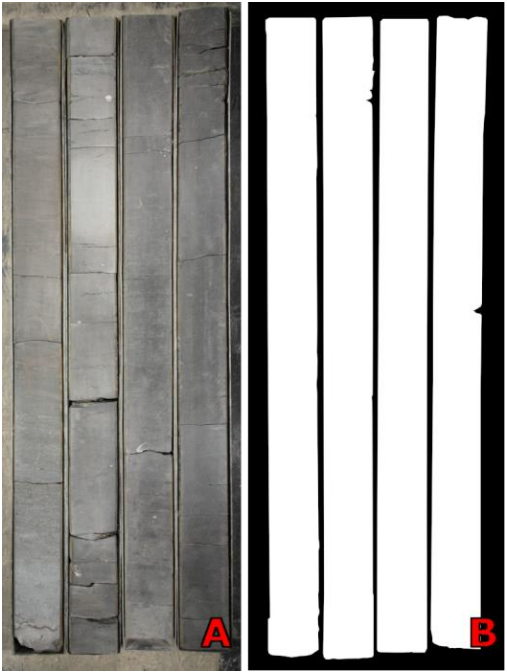

Figure 2. An example of a core box image (A) and its manually labelled mask (B). The mask is used for the training of a segmentation algorithm. The metrics are calculated based on the mask produced by the algorithm and the original mask

## 2.2. Algorithm selection

A fully convolutional artificial network was used for core detection as the core can have different shapes and appearances.

### 2.2.1. Network architecture

To have ease in deployment and enough model generalization, we chose U-Net. One of the most popular deep learning models for a semantic segmentation task is U-Net (Ronneberger et al., 2015), with ResNet-50 backbone available in the segmentation_models library above. The architecture can generalize data well, even within a small dataset (30-50 is sufficient for generalization of binary segmentation task in an invariant dataset). Compared with other models (such as PSPNet (Zhao et al., 2017), DeeplabV3 (Chen et al., 2017)), ResNet-50 shows the same metric values (discussed in the Results section).

## 2.3. Template-like augmentation

Augmentation is a standard tool to improve generalization by machine learning algorithms (DeVries and Taylor, 2017). The augmentation increases the dataset's size by adding different noises or changing a part of it. Various augmentations are available for images that can be applied in various scenarios. For example, the image can be distorted to simulate multiple shapes of an object or randomly cut-out to extract some features from an image. With the application of different augmentation techniques, a researcher can enhance algorithms performance (Shorten and Khoshgoftaar, 2019).

Unfortunately, due to large application areas, known libraries lack flexibility and are not suited for small datasets for a specific task. They are tuned to be applied during training and test periods (training time augmentation or TTA) to interchange an image with an augmented one. When an image is passed to the network, it could be randomly changed to an augmented one. It is not possible to fix the augmentation state for each training epoch. TTA is done to save disk space and provide large data variety. These augmentations can be saved, and these will be data permutations only, not new data. This work aims to use augmentations and create new data based on a researcher's data. We used known augmentation libraries to permute the data and developed a separate library to create new data based on the dataset we had already had.

A Python library was developed for template-like data augmentation (TLA). The library allows to interchange of segmented objects in an image to other objects of the same shape and semantic meaning from a collection of samples flexibly. The algorithm can change both: background and foreground parts of an image to increase the semantic segmentation algorithm performance on a new image., Various backgrounds can be used to cover the target object. An algorithm should ignore these backgrounds. A researcher can perform an adversarial attack using different objects. Thus, if there are several typical templates for the dataset (e.g., several store shelves, street views, etc.), it is possible to enlarge the dataset to an almost infinite number of images.

Background images were chosen from surrounding objects and made of images from a database by replacing the core with space from the background (figure 3). Some of these images were taken by a phone camera in different seasons and places to simulate situations where the core box lay on the ground. The size of the images was comparable to core box images (1908x4032 px).

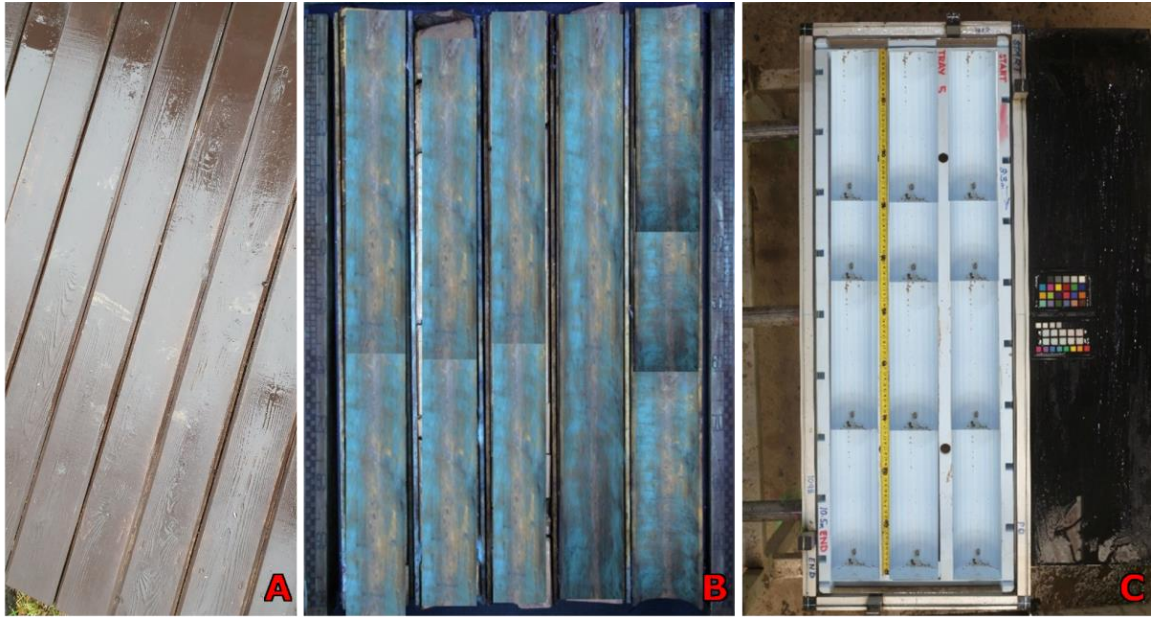

Figure 3. Core backgrounds examples. A- image from a phone camera, B and C – core box without a core.

### 2.3.1. Template-like augmentation (TLA) process

The augmentation algorithm uses simple image transformation techniques guided by an initial mask. There are several folders with identical to segmented labels names. An example of a folder structure is in figure 4:

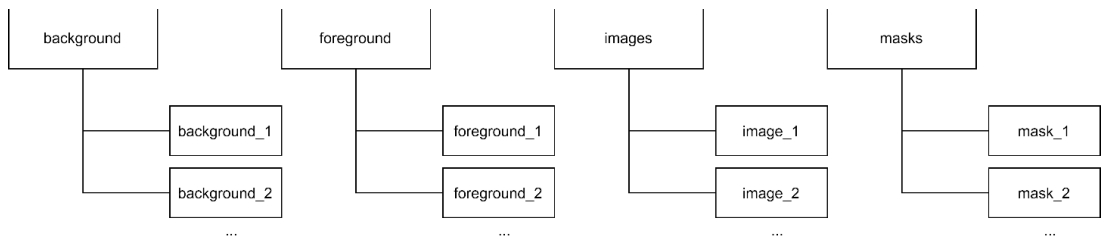

Figure 4. An example of the folder structure for the TLA augmentation process.

The initial mask and image are taken from a folder during augmentation (figure 5).

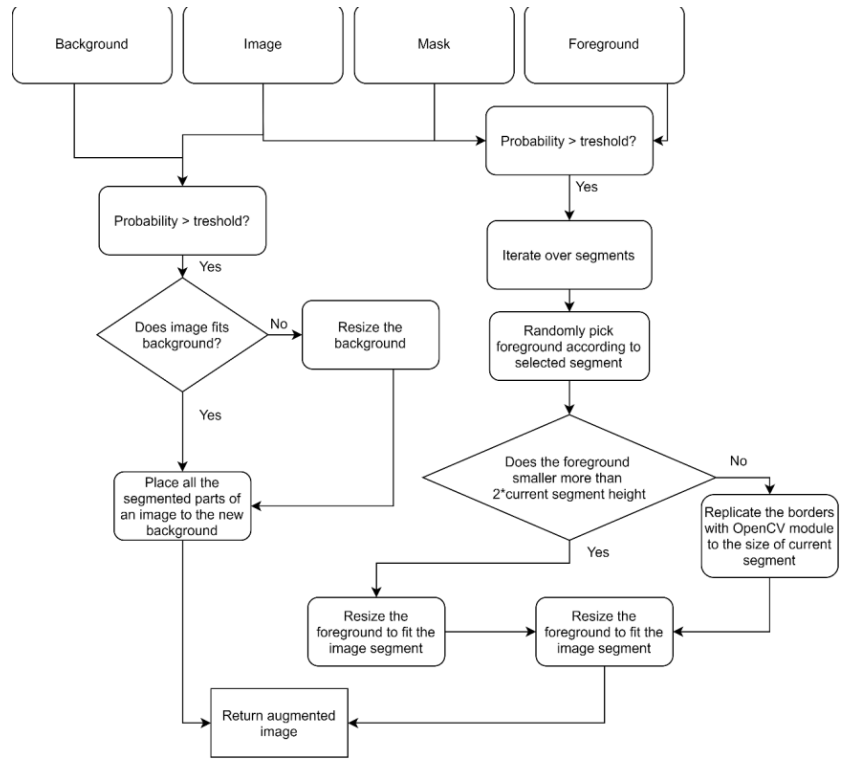

Figure 5. A scheme of the TLA augmentation process. The scheme is implemented in the TLA-augmentation algorithm available in GitHub.

A JSON file is loaded to match the class and mask with specific colour values (from 0 to 255 in greyscale). Depending on the probabilities set, different augmentations may be applied. Different parts of an image are chosen according to the mask, and specific segments (e.g., core or some other "foregrounds") are randomly picked from a "foreground" folder corresponding to the name matched with a mask in a JSON file. The segments are cropped to the size of the chosen part and replaced. A background may be randomly picked from the "background" folder that can be placed either on the top of an image or the bottom layer of an image in compliance with the settings set. The result of foreground augmentation is demonstrated in figure 6 and figure 7.

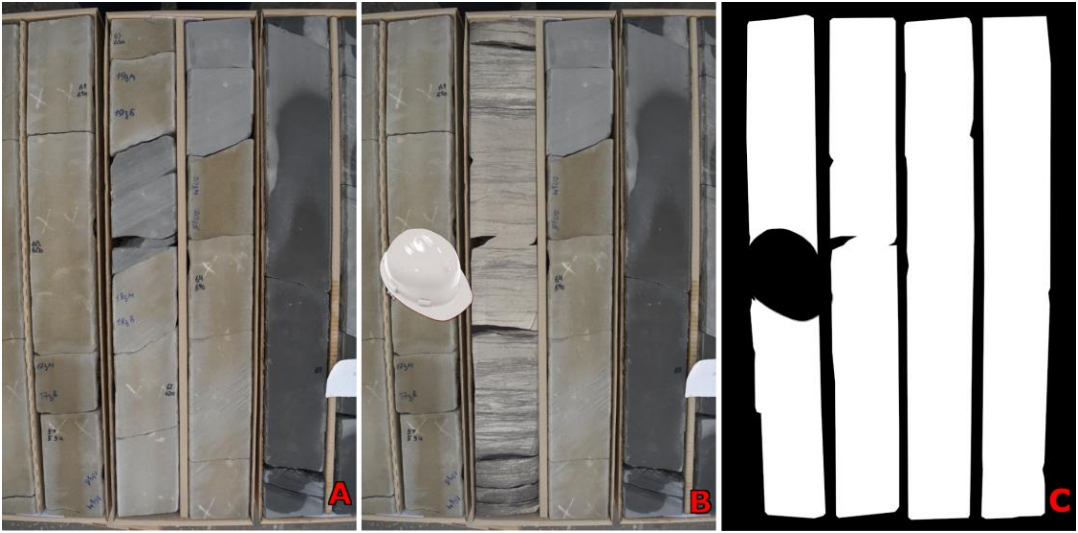

Figure 6. The results of foreground augmentation. A - original image, B - the image after augmentation, C - augmented mask,

Also, augmentations can be applied to masks and images by the Albumentations library to improve the efficiency of an algorithm. The obtained image sometimes may be confused with the initial image (figure 7).

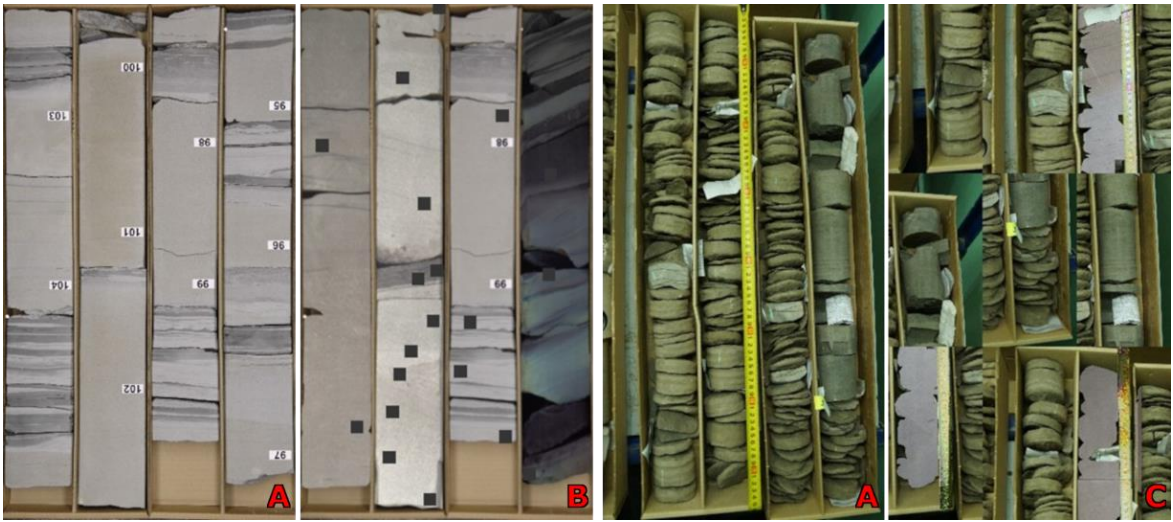

Figure 7. Comparison of the initial image (A) and augmented images (B, C). Some core in the core box changed (B); parts are cut-out (B) and mixed-up (C).

## 2.4. Data preprocessing

Firstly, we divided all images into a train set (734 images) and a test set (276). Secondly, we provided augmentation. Then we added extra 800 unlabeled images to the dataset from different data sources to compare the network's performance on a new set of data (with an application of template-like augmentation and without it). Each image was resized to a size of 320x480px to train a CNN. Images were normalized into 0 to 1 intervals using the min-max normalization before passing to the CNN.

## 2.5. Experiment setup

Three groups of data conditions were analyzed during experiments: a dataset without TLA augmentation (first setup, which included an augmentation from the Albumentation library), a composed dataset with TLA and non-augmented images (second setup), and a dataset with TLA only (third setup).

As the initial setup for neural network training, the hyperparameters were set to the following values: initial learning rate - 3e-4, batch size - 16, number of epochs - 15. The ReduceLROnPlateau was used as a scheduler to adjust the plateau's learning rate. The Early Stopping procedure was used to control the overfitting of the model. The complex loss function was used to train the model – binary cross-entropy and Jaccard loss with equal contribution to the optimization process.

Convolutional neural network training common practice includes data augmentation, which may be done with the Albumentations library. Following augmentations were used with each dataset group: resize, image flipping, image rotating, Gaussian noise and colour jitter.

The results were evaluated on a testing set with various metrics (Fawcett, 2006; Metz, 1978) and by a visual screening of new unlabeled images to apply an algorithm in the real world. A part of the visual screened results was also labelled to evaluate metrics performance. Pixel accuracy was not used in the semantic segmentation problems because it is not recommended (Rahman and Wang, 2016).

The metrics used are:

- $precision = \frac{TP}{TP+FP}$ (Fawcett, 2006; Metz, 1978) (1)

- $recall = \frac{TP}{P}$ (Fawcett, 2006; Metz, 1978) (2)

- f – score $F_\beta = \frac{(1+\beta^2)*precision*recall}{(\beta^2*precision)+recall}$ (Sasaki and Fellow, 2007) (3)

- intersection over union $IOU = \frac{TP}{(FP+TP+FN)}$ (Rahman and Wang, 2016) (4)

where TP – true positive, TN – true negative predictions, FP – false positive, FN – false negative labels to predict (the number of examples).

Precision (1) describes an amount of data predicted correctly out of all predicted values. Recall (2) describes the amount of data that is detected out of all data. The f1 score (3) means an equal contribution of precision and recall (β=1), f2 (3) means the inclination to recall (β=2). IOU(4) describes the prediction and target mask intersection. To illustrate the calculation, a demo-example was created (figure 8).

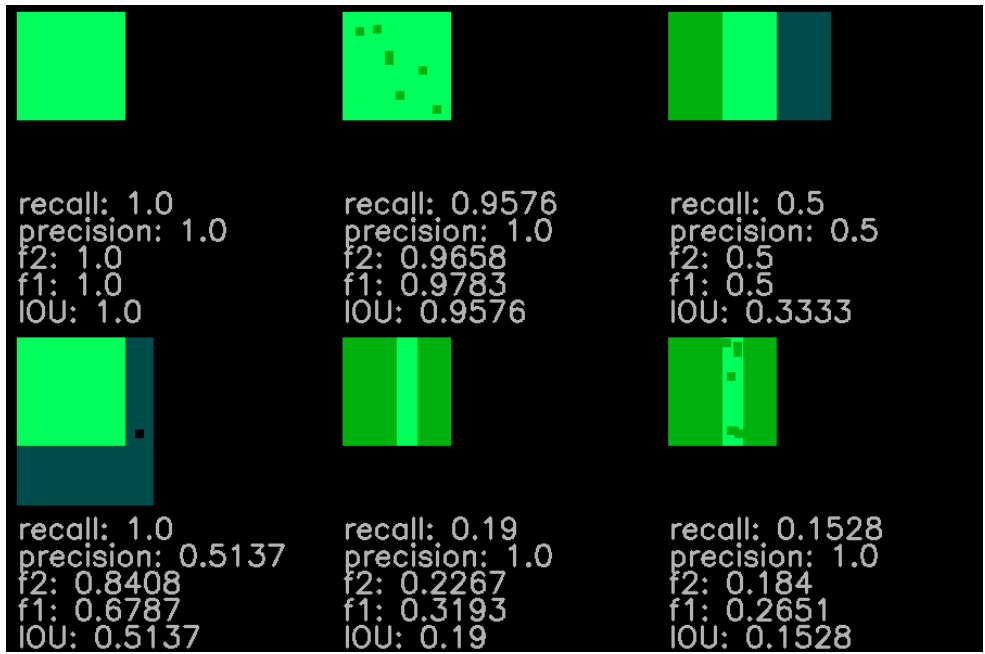

Figure 8. Examples of metrics calculation unrelated to core box segmentation problem. The ground truth is marked as green, the intersection with the mask is light green, the mask is light blue. The calculated metrics for different types of intersections are below the squares.

Convergence of the training loss and metrics values showed no overfitting or underfitting (figure 9) as train and validation curves are repeating each other behaviour. In the case of overfitting, the validation curve would go up for the loss, and the training curve would go down and vice-versa for IOU.

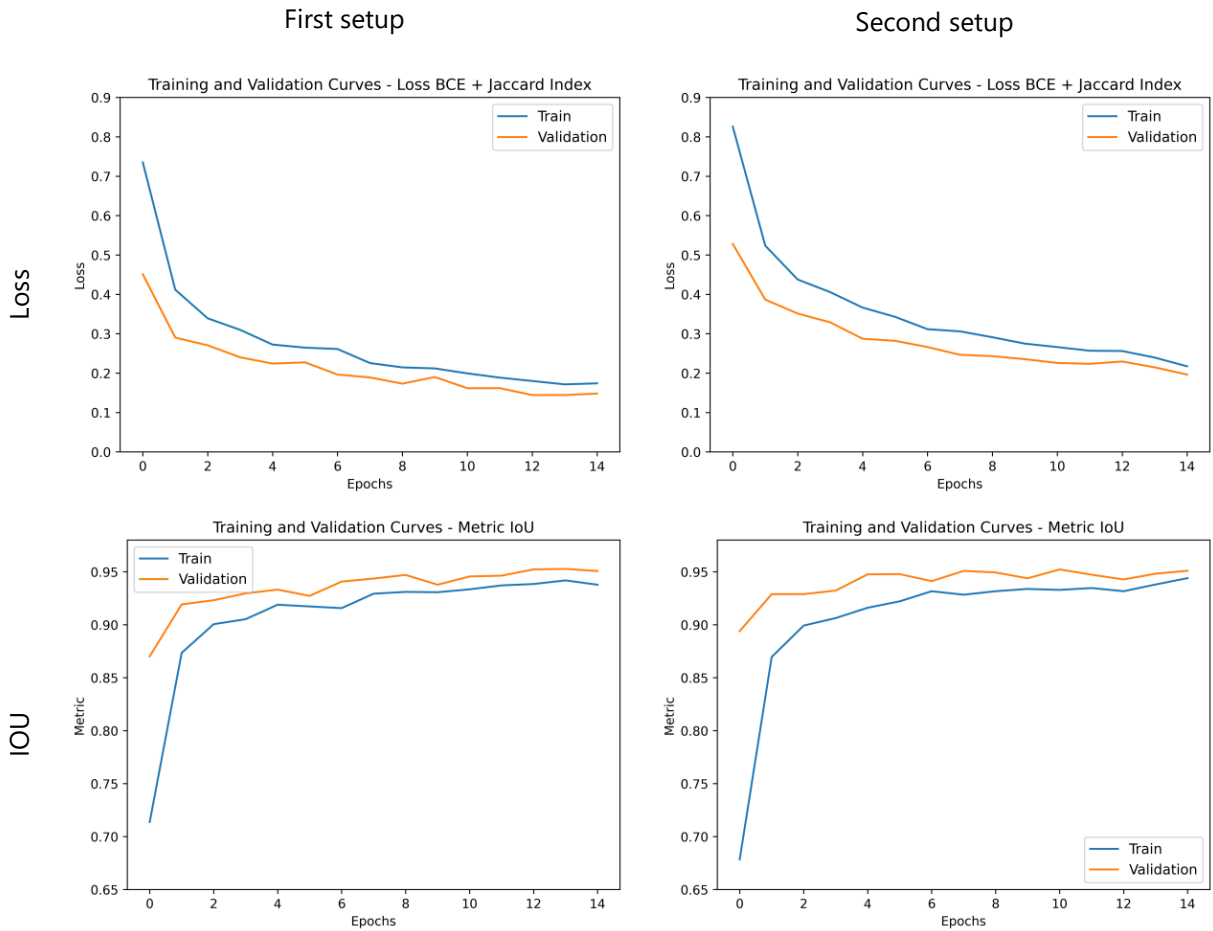

Figure 9. Loss and IOU curves for different data setups. First setup – the initial dataset separated by training and validation dataset, second setup – the training dataset expanded with TLA data.

### 3. Results and discussion

In the series of experiments, we discovered that the CNNs segment the images differently. The first setup training tended to segment most new images as a target value. The second setup training provided the best results during visual screening on any data, whether similar to the training set images or new images. The third setup worked only on a few images; thus, we compare here just two first settings (figure 10).

The prediction results were visually screened, and positive and negative cases were found. Better segmentation results determine positive cases with the second setup over the first setup.

Results received from new images predicted with both CNN's trained with and without TLA data were determined as positive and negative cases after the visual screening. Positive case - is the situation when the second setup CNN prediction results under visual control were more specific than the first setup CNN. The negative case – is an opposite situation when the results of the first setup CNN prediction under visual control were more specific than the second setup CNN.

| # | Original image | First setup CNN | Second setup CNN | Metrics w/o TLA | TLA |
|---|---|---|---|---|---|
| 1 | 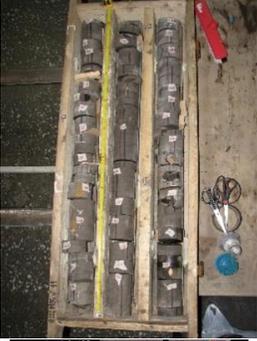 | 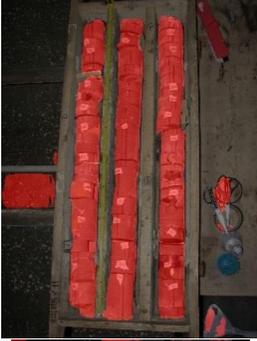 | 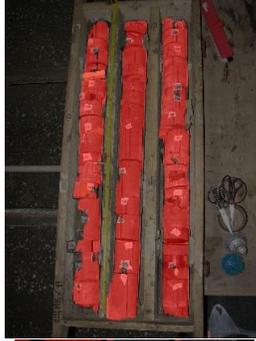 | IOU: 0,752 F1: 0,859 F2: 0,801 Recall: 0,976 Precision: 0,766 | IOU: 0,855 F1: 0,922 F2: 0,903 Recall: 0,955 Precision: 0,891 |
| 2 | 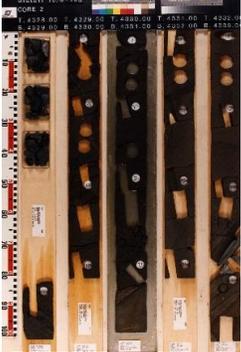 | 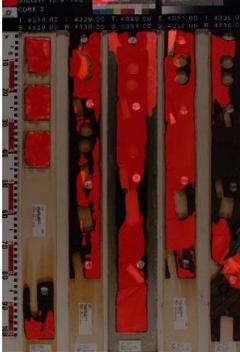 | 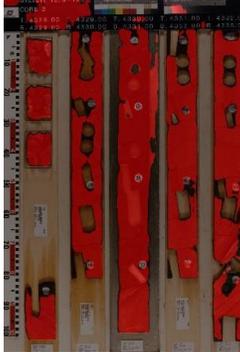 | IOU: 0,566 F1: 0,723 F2: 0,786 Recall: 0,637 Precision: 0,835 | IOU: 0,827 F1: 0,905 F2: 0,889 Recall: 0,933 Precision: 0,879 |
| 3 | 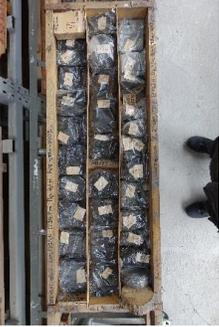 | 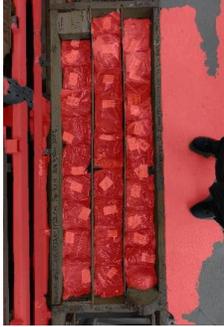 | 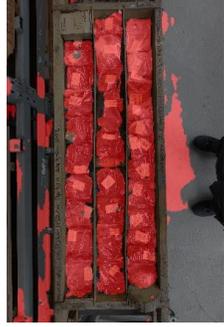 | IOU: 0,371 F1: 0,541 F2: 0,425 Recall: 0,993 Precision: 0,372 | IOU: 0,672 F1: 0,804 F2: 0,721 Recall: 0,992 Precision: 0,675 |
| 4 | 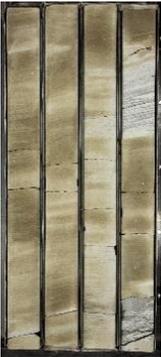 | 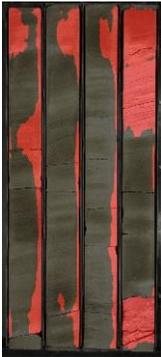 | 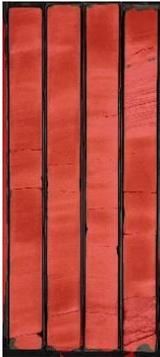 | IOU: 0,285 F1: 0,444 F2: 0,664 Recall: 0,286 Precision: 0,993 | IOU: 0,960 F1: 0,980 F2: 0,973 Recall: 0,992 Precision: 0,968 |

Figure 10. The comparison of an algorithm trained with two data setups. The rows contain different types of core boxes. 1,3- taken by hand with a camera, 2,4 - taken with a camera placed on a core-imaging stand.

The masks generated by the algorithm were manually edited to get the metrics. Here is the comparison table of different metrics calculated on both results in table 1.

Table 1. Metrics gained upon evaluation of both algorithms.

|  | **Regular Training** | | | | | **Training with TLA** | | | | |
| --- | --- | --- | --- | --- | --- | --- | --- | --- | --- | --- |
|  | **iou** | **f1** | **f2** | **precision** | **recall** | **iou** | **f1** | **f2** | **precision** | **recall** |
| mean | **0,826** | **0,900** | 0,890 | **0,925** | 0,886 | 0,801 | 0,883 | 0,890 | 0,880 | **0,897** |
| median | **0,838** | **0,912** | 0,913 | **0,956** | 0,916 | 0,815 | 0,898 | 0,913 | 0,916 | **0,919** |
| max | **0,999** | **1,000** | **1,000** | 0,999 | **1,000** | 0,984 | 0,992 | 0,995 | **1,000** | 0,999 |
| min | **0,371** | **0,541** | 0,425 | **0,504** | 0,372 | 0,256 | 0,407 | **0,523** | 0,298 | **0,612** |

First setup CNN slightly outperforms the second setup CNN (the delta is around 0.01-0.02), as seen on a table except for the recall parameter (the recall is slightly higher). The results mean that the TLA-based approach detects core 1,2% better but 4% less accurate. The minimum recall of the TLA-based approach is 1.6 higher than that of an algorithm trained with regular data (delta is 0.24). The TLA-based approach improved recall parameters. If a user uploads a new image, the overall performance will be measured by visual inspection of a result. The better the recall parameter, the better visual inspection results. Thus, our results improve the user experience.

### 3.1. Positive cases

As shown in figure 10, examples 3 and 4, the overall network performance accuracy increased on new data types. Specifically, the areas counted as "core" reduced in figure 10, images 1 and 3 and expanded in figure 10, 2 and 4. In the second setup, CNN started to mislabel some parts of the second image in figure 10, but the whole segmentation quality has grown to 1.4 by recall and IOU parameters.

The proposed method of template-like augmentation can help the algorithm determine which part of an image belongs to the core and which one is not (figure 11). In the initial dataset, any part of the core covered by sample label marks (e.g., a piece of paper with a number) was counted as "core" (covered with the same white mask as core). After training with template-like augmentation, the network distinguished such cases (figure 11).

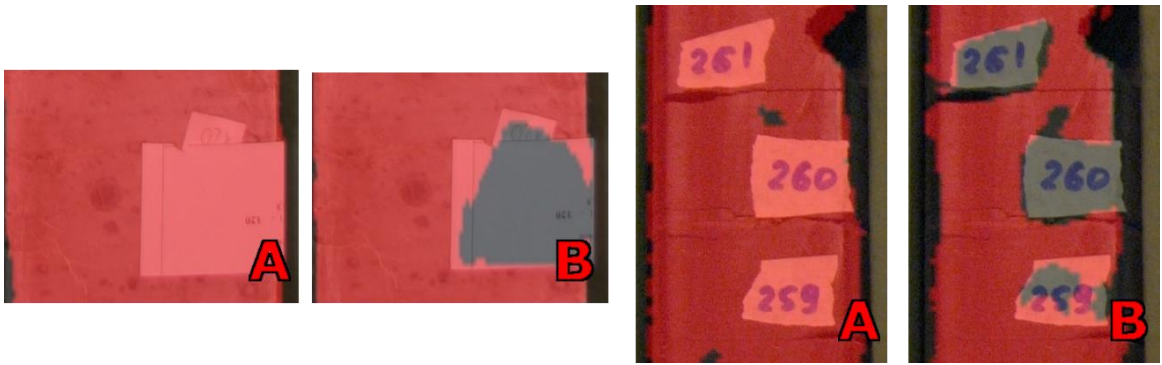

Figure 11. Comparison of results for regular training (A) and training using template-like augmentation (B).

### 3.2. Negative cases

Several types of negative cases were found (figure 12):

1) empty boxes were labelled as core (1% of cases).

2) areas around the core box were labelled as a core (3% of cases).

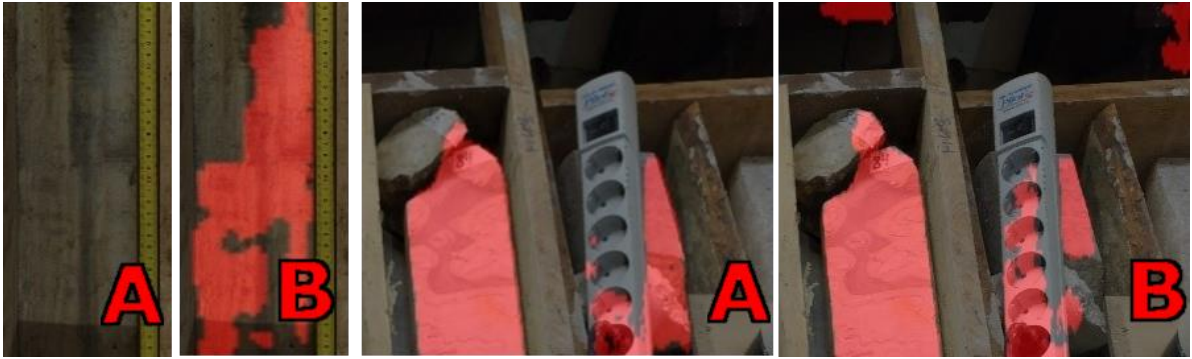

Figure 12. Comparison of new data segmentation results for regular training (A) and training using template-like augmentation (B).

### 3.3. Image understanding

A statistical approach was designed to filter out the "negative" and "positive" examples during images postprocessing to improve the algorithm's effectiveness. The bounding box approach implemented in the OpenCV library was used to extract the core from the box. It produced bounding boxes in the following way: the starting coordinates of a box – x and y and the bounding box's width and height. There are several characteristics of bounding boxes to look at. Firstly, the number of boxes in images varies from 1 to 6. This information was used to roughly evaluate the performance of an algorithm with simple bounding box counting. Secondly, the core can be found at the upper part of an image, so we can use this information to filter some boxes by the y coordinate. In combination with bounding box width and height, we can have several easy-to-estimate statistics that can increase the core extraction results for the user and further increase the recall (figure 13, figure 14).

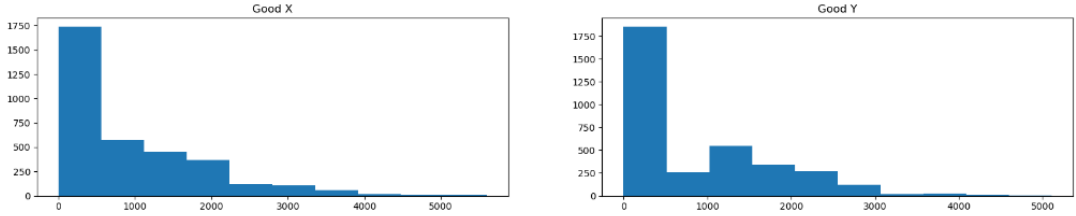
Figure 13. Distribution of x and y coordinates for good core bounding boxes detection examples.

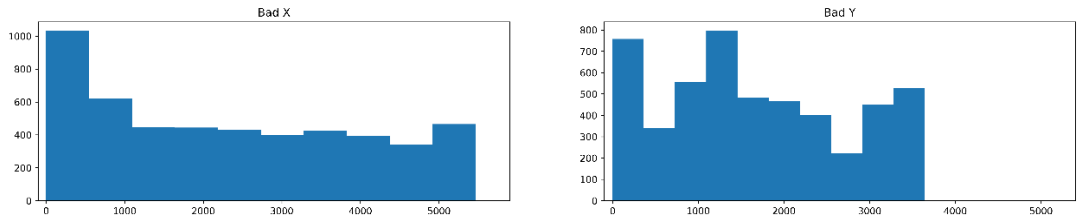
Figure 14. Distribution of x and y coordinates for "bad" core bounding boxes detection examples.

Based on statistical analysis, different cleaning-up approaches were developed:

1) based on a median width and height of a core.

   After mask generation, the bounding boxes detected with the OpenCV library and their respective median height and width are calculated. After that, the median can be used to calculate boundary conditions for filtering.

   xt_up = x * n; xt_bot = y/n,                                                              (6)

   where $n$ – a manually set coefficient (varies from 1.2 to 1.5), x – median width, y - median height, *xt_up* - the threshold for the upper boundary of a core mask width, *xt_bot* – the bottom boundary of width.

   The same algorithm can be applied for height.

   As a demonstration, there are 10 bounding boxes on an image that were detected. The calculated median width was 430, so after filtering, 5 core columns left as the width of the others were either lower or higher than the threshold set (from 358 (*xt_bot*) to 516 (*xt_up*), since coefficient n was equal to 1.2).

2) based on an image width threshold ratio (the core width cannot be less than this ratio).

   *gt = xi/m*,                                                                                    (5),

   where *gt* – global image width ratio, *xi* – the image width, *m* – manual set coefficient (varies from 50 to 200).

   For example, the image width is 4000, the *m* coefficient is 100, so the minimum core width is 400. Followed by the previous demonstration, the same columns can be filtered out and received 6 bounding boxes as they coincide with the minimum core width threshold (*gt*).

Both methods were used after experiments to increase the results for prediction and inform a user about the current segmentation status.

The recall was increased with the application of median size filters (Table 2).

Table 2. The metrics gained after postprocessing of results.

|  | Regular Training | | | | | Training with TLA | | | | |
| --- | --- | --- | --- | --- | --- | --- | --- | --- | --- | --- |
|  | iou | f1 | f2 | precision | recall | iou | f1 | f2 | precision | recall |
| mean | **0,813** | **0,892** | **0,893** | **0,900** | 0,896 | 0,777 | 0,868 | 0,887 | 0,846 | **0,904** |
| median | **0,827** | **0,906** | **0,916** | **0,930** | 0,923 | 0,797 | 0,887 | 0,907 | 0,889 | **0,929** |
| max | **0,976** | **0,988** | **0,995** | **0,989** | 1,000 | 0,970 | 0,985 | 0,991 | 0,981 | 1,000 |
| min | **0,382** | **0,553** | 0,438 | **0,490** | 0,385 | 0,239 | 0,386 | **0,511** | 0,274 | **0,610** |

### 3.4. Algorithm application and development

The developed algorithm can be applied widely to process core boxes images. The algorithm was used to extract separate columns out of a core-box image for further image processing. After extraction, the individual images can be used for automated rock typing or core quality analysis.

The developed algorithm can be set into a core description system as a web app for fast and easy core column extraction (figure 15). As mentioned, the calculated statistics may inform a user if some mistakes may appear on a mask. For example, if some samples marked as "core" are sufficiently larger or smaller than the median – the user can be informed to pay attention to such an image.

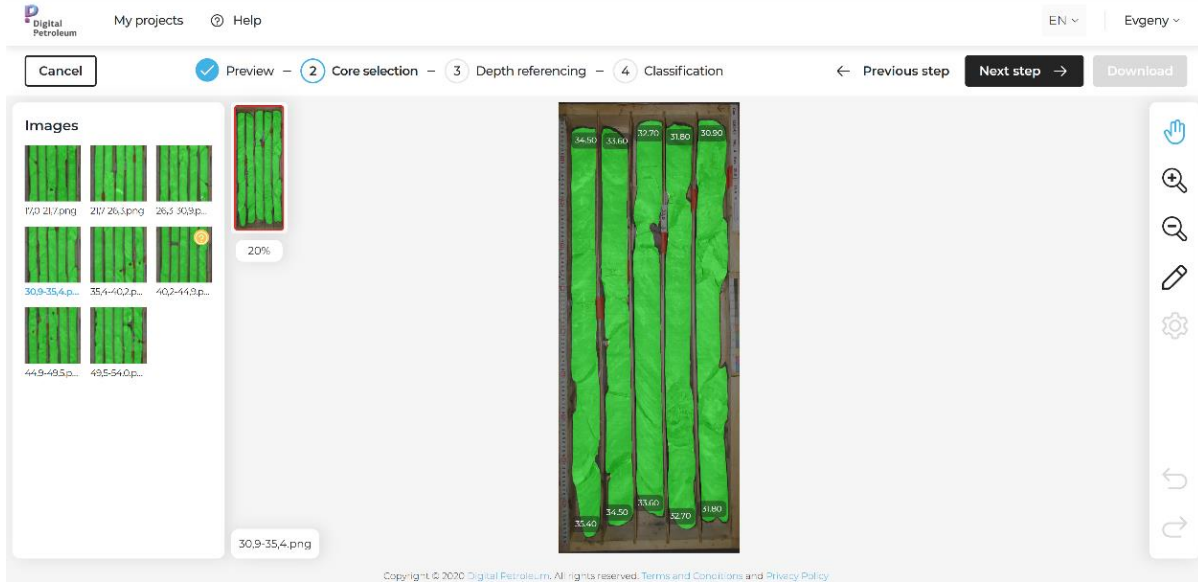

Figure 15. The algorithm implementation for production purposes. The algorithm produced the green mask, and a user can edit it.

The algorithm allowed to speed up the process of core extraction. With additional information (e.g., the depth and core layout direction, which can be provided by file name or user), core depth referencing can be auto-generated. After depth estimation user can interactively change the depth, correct the finite mask, and get extracted columns with depth referencing (figure 16).

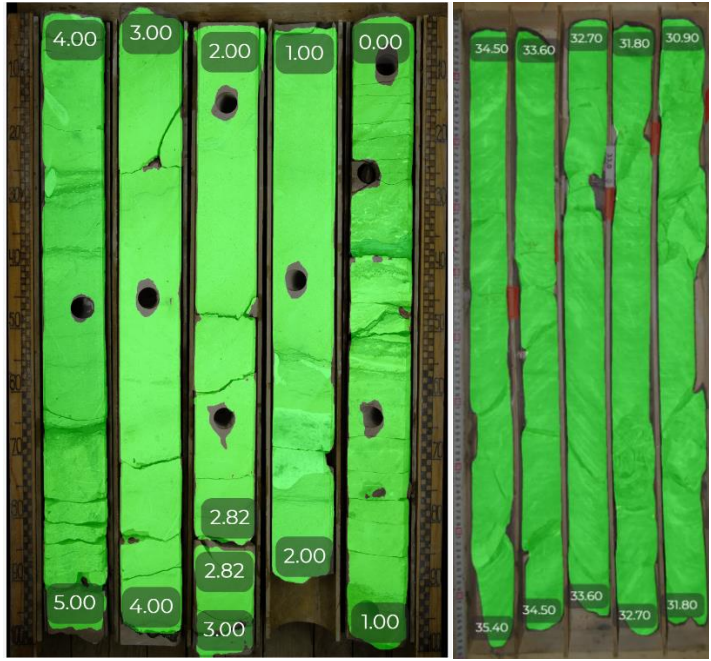

Figure 16. Example of image processing with information about depth and column height.

A user can get a significant speedup of core extraction. Manual core column extraction can take up to 20 minutes, depending on the number of core columns and depth referencing an image. Automated core column extraction takes only one minute. It is independent of depth referencing and the number of columns—the application of the developed algorithm speedups core extraction to 20x for a user. Days of manual core extraction turn out to be several minutes with the application of the algorithm.

**4. Conclusions**

The work firstly presented a method to extract core columns from a full-bore core image.

The new data augmentation method helps create new data when samples quantity is low for generating datasets with GANs. In this research, the method was applied to core box images segmentation.

Several setups were used to compare the performance of U-Net-like convolutional neural networks (CNN) training approaches. The first setup included traditionally prepared data, and the second had the same data with the addition of TLA-generated data. All setups were tested on a new bunch of data used neither on training nor on validation steps.

Comparing the performance of the convolutional neural network (CNN), several conclusions were made. Firstly, the network trained with the TLA method increased recall. Secondly, TLA-generated data improved the segmentation results on new data not presented to the segmentation algorithm and sufficiently distinct from examples provided during the training and validation period. Thirdly, the CNN trained with TLA-data started to distinguish the areas mislabeled as target label "core" on the training dataset.

The results of statistical analysis of masks generated by the algorithm allow one to post-process the mask, clean up final results, and inform a user about mistakes presented in images.

The trained algorithm was set into the system for automated core description to generate the automatic description or export images to external software programs with a speedup of a single image processing time to 20x. A user can use the previously spent time for core image extraction for much more exciting tasks.

The trained algorithm can analyze provided information about the depth of a core box and include this information in its description.

## 5. Funding

This research did not receive any specific grant from funding agencies in the public, commercial, or not-for-profit sectors.

## 6. Acknowledgements

We thank our colleagues from ZAO MIMGO, OOO IGT group, and Lomonosov MSU for consultancy and feedback on the algorithm performance during the paper preparation.

## 7. Roles of the authors

E.E. Baraboshkin – ideas and their implementation, design of experiments, data analysis, drafting.

A.E. Demidov – design and implementation of experiments, data analysis, drafting.

D.M. Orlov – ideas, design of experiments, drafting.

D.A. Koroteev – academic advisor, ideas, design of experiments, drafting.

## 8. Computer Code Availability

GitHub was used to place the augmentation pipeline as Open-Source code for TLA at https://github.com/BEEugene/TemplateArtification/.

## 9. Data Availability

Datasets related to this article can be found at open-source online data repositories: https://www.iodp.org/resources/access-data-and-samples IODP (Integrated Ocean Drilling Program), http://www.bgs.ac.uk/opengeoscience/photos.html (hosted by the British Geological Survey), http://geocollections.info/ (National Geological Collection of Estonia) and https://nopims.dmp.wa.gov.au/Nopims/ (Geoscience Australia), Volve dataset (Equinor).

## 10. References

Alzubaidi, F., Mostaghimi, P., Swietojanski, P., Clark, S.R., Armstrong, R.T., 2021. Automated lithology classification from drill core images using convolutional neural networks. J. Pet. Sci. Eng. 197, 107933. https://doi.org/10.1016/j.petrol.2020.107933

Baraboshkin, E.E., Ismailova, L.S., Orlov, D.M., Zhukovskaya, E.A., Kalmykov, G.A., Khotylev, O. V., Baraboshkin, E.Y., Koroteev, D.A., 2020. Deep convolutions for in-depth automated rock typing. Comput. Geosci. 135, 104330. https://doi.org/10.1016/j.cageo.2019.104330

Bradski, G., 2000. The OpenCV Library. Dr. Dobb's J. Softw. Tools 120, 122–125.

Bray, T., 2014. The JavaScript Object Notation (JSON) Data Interchange Format, RFC. https://doi.org/10.17487/rfc7158

Buslaev, A., Parinov, A., Khvedchenya, E., Iglovikov, V.I., Kalinin, A.A., 2018. Albumentations: fast and flexible image augmentations. ArXiv e-prints 1–4.

Chen, L.-C., Papandreou, G., Schroff, F., Adam, H., 2017. Rethinking Atrous Convolution for Semantic Image Segmentation.

Chollet, F., Google, Microsoft, Others, 2015. Keras [WWW Document]. https://keras.io.

DeVries, T., Taylor, G.W., 2017. Improved Regularization of Convolutional Neural Networks with Cutout. https://doi.org/10.1016/j.neuron.2007.06.026

Egorov, D., 2019. Extraction of Petrophysical Information and Formation Heterogeneity Estimation from

Core Photographs by Clustering Algorithms. https://doi.org/10.3997/2214-4609.201902190

Fawcett, T., 2006. An introduction to ROC analysis. Pattern Recognit. Lett. 27, 861–874. https://doi.org/10.1016/j.patrec.2005.10.010

Gong, R., Li, W., Chen, Y., Van Gool, L., 2018. DLOW: Domain Flow for Adaptation and Generalization.

Hunter, J.D., 2007. Matplotlib: A 2D Graphics Environment. Comput. Sci. Eng. 9, 90–95. https://doi.org/10.1109/MCSE.2007.55

Ketkar, N., 2017. Introduction to PyTorch, in: Deep Learning with Python. Apress, Berkeley, CA, pp. 195–208. https://doi.org/10.1007/978-1-4842-2766-4_12

Khasanov, I.I., Ponomarev, I.A., Postnikov, A.V., Osintseva, N.A., 2016. A method for capacity characteristics for pay rocks quantitative estimation with the application of digital core images processing (in Russian). Geomodel-2016.

Lepistö, L., 2005. Rock image classification using color features in Gabor space. J. Electron. Imaging 14, 040503. https://doi.org/10.1117/1.2149872

Martín Abadi, Ashish Agarwal, Paul Barham, E.B., Zhifeng Chen, Craig Citro, Greg S. Corrado, A.D., Jeffrey Dean, Matthieu Devin, Sanjay Ghemawat, I.G., Andrew Harp, Geoffrey Irving, Michael Isard, Rafal Jozefowicz, Y.J., Lukasz Kaiser, Manjunath Kudlur, Josh Levenberg, Dan Mané, M.S., Rajat Monga, Sherry Moore, Derek Murray, Chris Olah, J.S., Benoit Steiner, Ilya Sutskever, Kunal Talwar, P.T., Vincent Vanhoucke, Vijay Vasudevan, F.V., Oriol Vinyals, Pete Warden, Martin Wattenberg, M.W., Yu, Y., Zheng, X., 2015. TensorFlow: Large-scale machine learning on heterogeneous systems. Softw. available from

tensorflow.org.

Metz, C.E., 1978. Basic Principles of ROC analysis. Semin. Nucl. Med. 8, 283–298. https://doi.org/10.1016/S0001-2998(78)80014-2

Patel, A.K., Chatterjee, S., Gorai, A.K., 2017. Development of online machine vision system using support vector regression (SVR) algorithm for grade prediction of iron ores, in: 2017 Fifteenth IAPR International Conference on Machine Vision Applications (MVA). IEEE, pp. 149–152. https://doi.org/10.23919/MVA.2017.7986823

Pedregosa, F., Varoquaux, G., Gramfort, A., Michel, V., Thirion, B., Grisel, O., Blondel, M., Prettenhofer, P., Weiss, R., Dubourg, V., Vanderplas, J., Passos, A., Cournapeau, D., Brucher, M., Perrot, M., Duchesnay, É., 2011. Scikit-learn: Machine Learning in Python. J. Mach. Learn. Res. 12, 2825–2830.

Prince, C.M., Chitale, J., 2008. Core Image Analysis: Reliable Pay Estimation in Thin-Bedded Reservoir Units. Image Process. 1–6.

Rahman, M.A., Wang, Y., 2016. Optimizing intersection-over-union in deep neural networks for image segmentation, in: Lecture Notes in Computer Science (Including Subseries Lecture Notes in Artificial Intelligence and Lecture Notes in Bioinformatics). Springer Verlag, pp. 234–244. https://doi.org/10.1007/978-3-319-50835-1_22

Ronneberger, O., Fischer, P., Brox, T., 2015. U-Net: Convolutional Networks for Biomedical Image Segmentation.

Sasaki, Y., Fellow, R., 2007. The truth of the F-measure.

Shorten, C., Khoshgoftaar, T.M., 2019. A survey on Image Data Augmentation for Deep Learning. J. Big Data 6, 60. https://doi.org/10.1186/s40537-019-0197-0

Thomas, A., Rider, M., Curtis, A., MacArthur, A., 2011. Automated lithology extraction from core photographs. First Break 29, 103–109.

Travis, O.E., 2006. A guide to NumPy. Trelgol Publishing, USA.

Van Rossum, G., Drake, F.L., 2011. The Python language reference manual : for Python version 3.2. Network Theory Ltd.

Wieling, I.S., 2013. Facies and permeability prediction based on analysis of core images.

Yakubovskiy, P., 2019. Segmentation Models. GitHub Repos.

Zhao, H., Shi, J., Qi, X., Wang, X., Jia, J., 2017. Pyramid Scene Parsing Network.